\documentclass[letterpaper, 10 pt, conference]{ieeeconf}  

\IEEEoverridecommandlockouts                              
\usepackage{graphicx} 
\usepackage{amsmath} 
\usepackage{amssymb}  
\usepackage{hyperref}
\usepackage[table]{xcolor}

\DeclareMathOperator{\EX}{\mathbb{E}}

\title{\LARGE \bf
Sparsity in Variational Autoencoders
}

\author{Andrea Asperti
\thanks{A. Asperti is Professor at the University  of Bologna, Department of Informatics: Science and Engineering (DISI)}
}

\begin{document}

\maketitle
\thispagestyle{empty}
\pagestyle{empty}

\begin{abstract}
Working in high-dimensional latent spaces, the internal encoding of data in Variational
Autoencoders becomes naturally {\em sparse}. We discuss this known but controversial 
phenomenon sometimes refereed to as overpruning, to emphasize the under-use of the model capacity. In fact,
it is an important form of self-regularization, with all the typical benefits associated
with sparsity: it forces the model to focus on the really important features, highly
reducing the risk of overfitting. Especially, it is a major methodological guide for
the correct tuning of the model capacity, progressively augmenting it to attain sparsity, 
or conversely reducing the dimension of the network removing links to zeroed out neurons.
The degree of sparsity crucially depends on the network architecture: for instance, convolutional networks typically show less sparsity, likely due to the tighter relation of 
features to different spatial regions of the input. 
\end{abstract}

\section{Introduction}
Variational Autoencoders (VAE) (\cite{VAE13,RezendeMW14}) are a fascinating facet of autoencoders,
supporting, among other things, random generation of new data samples. Many interesting researches
have been recently devoted to this subject, aiming either to extend the paradigm, 
such as conditional VAE (\cite{CVAE-15,CVAE-16}), or to improve some aspects it, as in the case of 
importance weighted autoencoders (IWAE) and their variants (\cite{BurdaGS15,RainforthKLMIWT18}). From the 
point of view of applications, variational autoencoders proved to be successful for generating many 
kinds of complex data such as natural images \cite{GregorDGW15} or facial expressions \cite{YehLGA16}, or also, more interestingly, 
for making probabilistic predictions about the future from static images \cite{WalkerDGH16}. In particular, variational autoencoders 
are a key component of the impressive
work by DeepMind on representation and rendering of three-dimensional scenes, 
recently published on Science \cite{Eslami18}.

Variational Autoencoders have a very nice mathematical theory, that we shall briefly
survey in Section~\ref{sec:VAE}. One important component of the objective function
neatly resulting from this theory is the Kullback-Leibler divergence $KL(Q(z|X)||P(z))$, 
where $Q(z|X)$ is the distribution of latent variables $z$ given the data $X$ guessed
by the network, and $P(z)$ is a prior distribution of latent variables (typically, 
a Normal distribution). This component is acting as a regularizer, inducing a better
distribution of latent variables, essential for generative sampling.

An additional effect of the Kullback-Leibler component is that, 
working in latent spaces of sufficiently high-dimension, the network learns representations
sensibly more compact than the actual network capacity: many latent variables are 
zeroed-out independently from the input, and are completely neglected by the generator.
In the case of VAE, this phenomenon was first observed in \cite{BurdaGS15}; following a 
terminology introduced in \cite{overpruning17}, it is frequently referred to as overpruning, to stress
the fact that the model is induced to learn a suboptimal generative model by limiting itself to exploit a small number of latent variables. From this  point of view, it is usually regarded as a negative property, and different training mechanisms have been envisaged to
tackle this issue (see Section~\ref{sec:tackling}). In this article, we take a slightly different
perspective, looking at {\em sparsity} of latent variables as an important form
of self-regularization, with all the typical benefits associated with it: in
particular, it forces the model to focus on the really important features, typically
resulting in a more robust encoding, less prone to overfitting. Sparsity is usually
achieved in Neural Network by means of weight-decay L1 regularizers (see e.g. \cite{DeepBook}), 
and it is hence a pleasant surprise to discover that a similar
effect is induced in VAEs by the Kullback-Leibler component of the objective function.

The most interesting consequence is that, for a given architecture, there seems 
to exist an {\em intrinsic} internal dimension of data. This property can be exploited 
as a main methodological guideline to tune the network capacity, 
progressively augmenting it to attain sparsity, or conversely reducing the dimension of the network removing 
links to unused neurons. 

The degree of sparsity depends on the network architecture: for instance, convolutional networks typically show less sparsity, likely due to the tighter relation of 
features to different spatial regions of the input. This seems to suggest that the
natural way to tackle the sparsity phenomenon is not by means of hand-tuned schemes
that may just remove the regularization effect of the Kullback-Leibler term, but
by more sophisticated networks, inducing a less correlated use of latent variables
(as e.g. in ).

\section{Variational Autoencoders}\label{sec:VAE}
In this section we briefly recall the theory behind variational autoencoders;
see e.g. \cite{tutorial-VAE} for a more thoroughly introduction. 

In latent variable
models we express the probability of a data point $X$ through marginalization over
a vector of latent variables:
\begin{equation}P(X) = \int P(X|z,\theta)P(z)dz \label{eq1} \approx \EX_{z\sim P(z)}P(X|z,\theta)
\end{equation}
where $\theta$ are parameters of the model (we shall omit them in the sequel).

Sampling in the latent space may be problematic for several reasons. The
variational approach exploits sampling from an auxiliary distribution $Q(z)$.
In order to understand the relation between $P(X)$ and $\EX_{z\sim Q(z)}P(X|z,\theta)$ 
it is convenient to start from  the Kullback-Leibler divergence
of $Q(z)$ from $P(z|X)$:
\begin{equation}
    KL(Q(z)||P(z|X)) = \EX_{z\sim Q(z)}log\frac{Q(z)}{P(z|X)}
\end{equation}
or also, exploiting Bayes rule,
\begin{equation}
    KL(Q(z)||P(z|X)) = \EX_{z\sim Q(z)}log\frac{Q(z)P(X)}{P(X|z)P(z)}
\end{equation}
$P(X)$ does not depend on $z$ and may come out of the expectation; rephrasing
part of the right hand side in terms of the KL divergence of $Q(z)$ from $P(z)$
we obtain, by simple manipulations:
\begin{equation}
\begin{array}{l}
    log(P(X)) - KL(Q(z)||P(z|X)) = \\
    \hspace{1cm} \EX_{z\sim Q(z)}log(P(X|z)) - KL(Q(z)||P(z))\label{eq:elbo}
    \end{array}
\end{equation}
Since the Kullback-Leibler divergence is always positive, the term on
the right is a lower bound to the loglikelihood $P(X)$, known as Evidence Lower Bound (ELBO).

In Equation \ref{eq:elbo}, $Q(z)$ can be any distribution; in particular, 
we may take one depending on $X$, hopefully resembling $P(z|X)$ so that 
the quantity $KL(Q(z)||P(z|X))$ is small; in this case the loglikelihood
$P(X)$ is close to the Evidence Lower Bound; our learning
objective is its maximization:
\begin{equation}
\begin{array}{l}
    log(P(X)) \approx \\
    \hspace{.5cm}\EX_{z\sim Q(z|X)}log(P(X|z) - KL(Q(z|X)||P(z))
\end{array}\label{eq:obj}
\end{equation}
The term on the right has a form resembling an autoencoder, where
the term $Q(z|X)$ maps the input $X$ to the latent representation $z$, 
and $P(X|z)$ decodes $z$ back to $X$. 

The common assumption in variational
autoencoders is that $Q(z|X)$ is normally distributed around an encoding
function $\mu_\theta(X)$, with variance $\sigma_\theta(X)$; similarly 
$P(X|z)$ is normally distributed around a decoder function $d_\theta(z)$.
All functions $\mu_\theta$, $\sigma_\theta$ and $d_\theta$ are computed
by neural networks. 

Provided the decoder function $d_\theta(z)$ has enough power, the shape of the prior distribution $P(z)$ for latent variables can be arbitrary, and
for simplicity we may assume it is a normal distribution
\[P(z) = G(0,1)\]
The term $KL(Q(z|X)||P(z)$ is hence the KL-divergence between two Gaussian distributions $G(\mu_\theta(X),\sigma^2_\theta(X))$ and $G(1,0)$ which can be 
computed in closed form:
\begin{equation}\label{eq:closed-form}
\begin{array}{l}
    KL(G(\mu_\theta(X),\sigma_\theta(X)),G(0,1)) = \\
    \hspace{1cm}\frac{1}{2}(\mu_\theta(X)^2 + \sigma^2_\theta(X)-log(\sigma^2_\theta(X)) -1)
\end{array}
\end{equation}
As for the term $\EX_{z\sim Q(z|X)}log(P(X|z)$, under the Gaussian assumption the logarithm of $P(X|z)$ is just the quadratic distance between $X$ and 
its reconstruction $d_\theta(z)$. Sampling, according to $Q(z|X))$ is easy, since we know the moments $\mu_\theta(X)$ and $\sigma^2_\theta(X)$ of this
Gaussian distribution. The only remaining problem is to integrate sampling 
with backpropagation, that is solved by the well known reparametrization trick (\cite{VAE13,RezendeMW14}).

Let us finally observe that, for generating new samples, the mean and variance
of latent variables {\em is not used}: we simply sample a vector of latent
variables from the normal distribution $G(0,1)$ and pass it as input
to the decoder. 

\section{Discussion}\label{sec:discussion}
The mathematical theory behind Variational Autoencoders is very neat;
nevertheless, there are several aspects whose practical relevance is difficult to grasp and look almost counter-intuitive. For instance, 
in his Keras blog on Variational Autoencoders\footnote{\href{https://blog.keras.io/building-autoencoders-in-keras.html}{https://blog.keras.io/building-autoencoders-in-keras.html}}, F.Chollet writes - talking about the Kullback-Leibler component in the objective function - that ``you can actually get rid
of this latter term entirely''. In fact, getting rid of it, the Gaussian distribution of latent variables would tend to collapse to 
a Dirac distribution around its mean value, making sampling pointless:
the variational autoencoder would resemble a traditional autoencoder,
preventing any sensible generation of new samples from the latent space.

Still, the relevance of sampling during the training phase, apart from reducing overfitting and improving the robustness of the autoencoder, 
is not so evident. The variance $\sigma^2_\theta(X)$ around the encoding $\mu_\theta(X)$ is typically very small, reflecting the
fact that only a small region of the latent space will be able to
produce a reconstruction close to $X$. Experimentally, we see that
the variance decreases very quickly during the first stages of
training; since we rapidly reduce to sample in a small
area around $\mu_\theta(X)$, it is natural to wonder about the
actual purpose of this operation.

Moreover, the quadratic penalty on
$\mu_\theta(X)$ in Equation~\ref{eq:closed-form} is already sufficient to induce a Gaussian-like distribution of latent variables: so
why we try to keep variance close to $1$ if we reasonably expect it
to be much smaller?  

In the following sections we shall try to give some empirical answers to
these questions, investigating encodings for different datasets, using different
neural architectures, with latent spaces of growing dimension. This will lead
us to face the interesting sparsity phenomenon, that we shall discuss
in Section~\ref{sec:sparsity}.

\section{MNIST}

\begin{figure}[ht]
    \centering\vspace{-.46cm}
    \includegraphics[width=.45\textwidth]{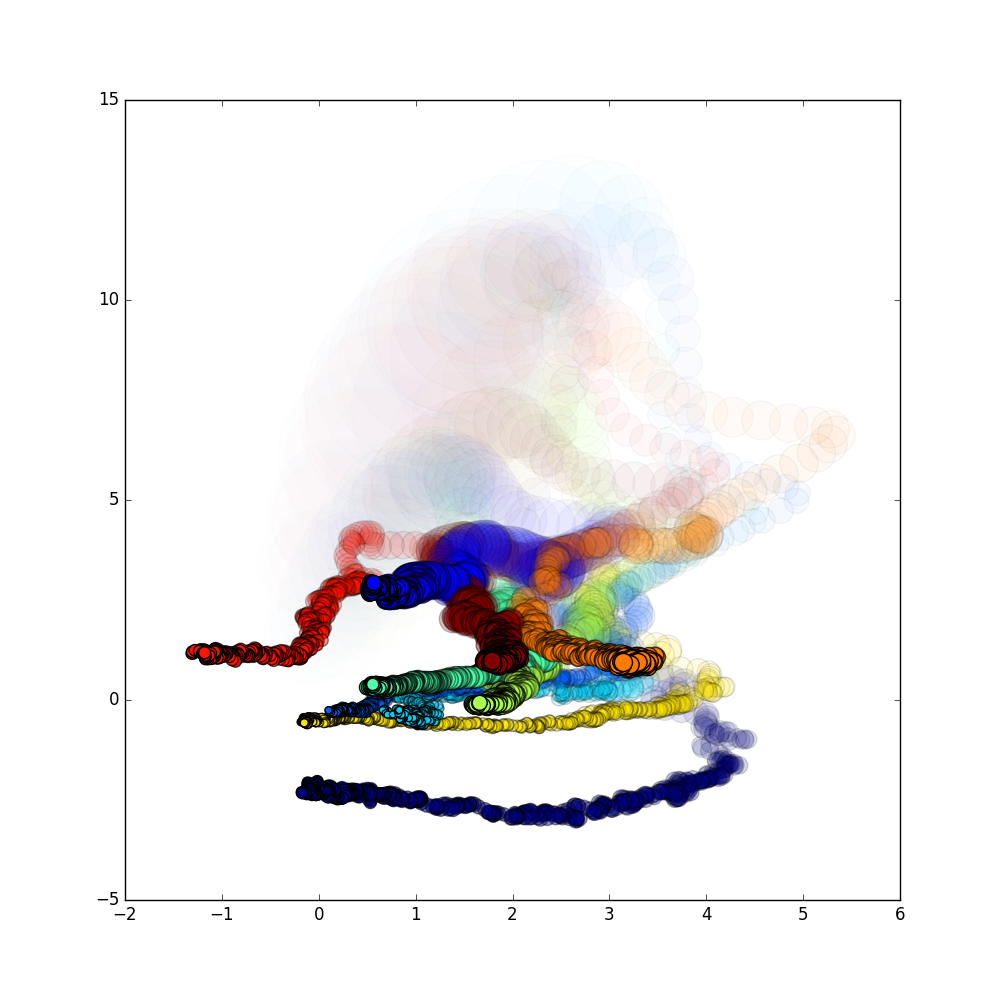}\vspace{-.3cm}
    \caption{Trajectories of ten MNIST digist in a binary latent space during the first epoch of training; pictures fade away with time. The area of the circle is proportional to the variance: it is initally close to 1, but it
    rapidly decrease to very small values.}
    \label{fig:fading}
\end{figure}

In a video available on line\footnote{\href{http://www.cs.unibo.it/~asperti/variational.html}{http://www.cs.unibo.it/\textasciitilde{}asperti/variational.html}}
we describe the trajectories in a binary latent space followed by ten random digits of the MNIST dataset (one for each class) during
the first epoch of training. The animation is summarized in 
Figure\ref{fig:fading}, where we use fading to describe the evolution in time.

\begin{figure}[ht]
    \centering\vspace{-.46cm}
    \includegraphics[width=.45\textwidth]{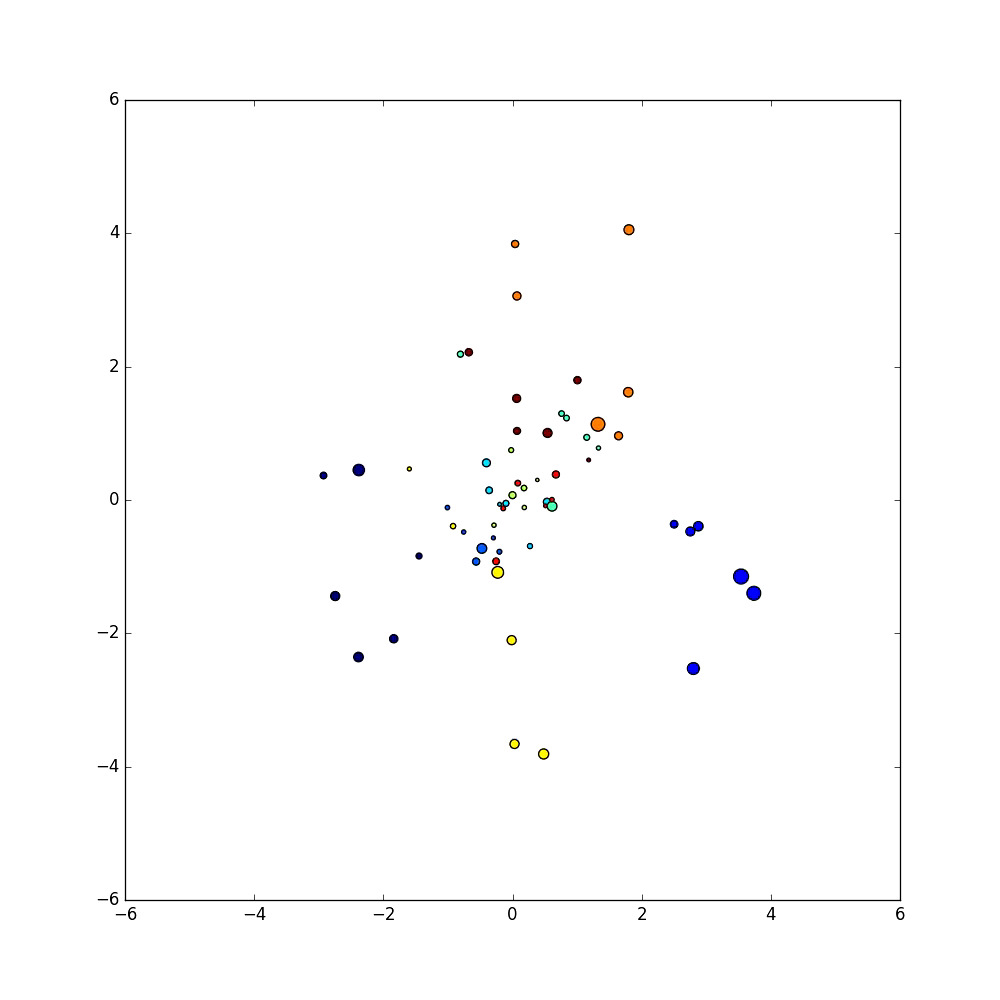}\vspace{-.3cm}
    \caption{Position and variance of 60 MNIST digits after 10 epochs of training. Observe the Gaussian-like distribution and, especially, the really small values of the variance}
    \label{fig:Gaussian_distribution}
\end{figure}

Each digit is depicted by a circle with an area proportional to its
variance. Intuitively, you can think of this area as the portion of the latent
space producing a reconstruction similar to the original.
At start time, the variance is close to 1, but it rapidly gets much
smaller. 
This is not surprising, since we need to find a place 
for 60000 {\em different} digits.
Note also that
the ten digits initially have a chaotic distribution, but progressively dispose themselves
around the origin in a Gaussian-like shape. 
This Gaussian distribution is better appreciated in Figure \ref{fig:Gaussian_distribution}, where we describe the position and ``size'' (variance) of 60 digits (6 for each class) after 10 training epochs.

The real purpose of sampling during training is to induce the generator to exploit 
{\em as much as possible} the latent space surrounding the encoding of each data point. 
How much space can we occupy? In principle, {\em all the available space}, that is why we try to keep the distribution $P(z|y)$ close to a normal distribution (the entire latent space). 
The hope is that this should induce a better coverage of the latent space, resulting in a better generation of new samples.

In the case of MNIST, we start getting some significant empirical evidence of the previous fact when considering a sufficiently deep architecture in a latent space of dimension 3 (with 2 dimensions it
is difficult to appreciate the difference).

\begin{figure}[ht!]
    \begin{tabular}{c}
    \includegraphics[width=.4\textwidth]{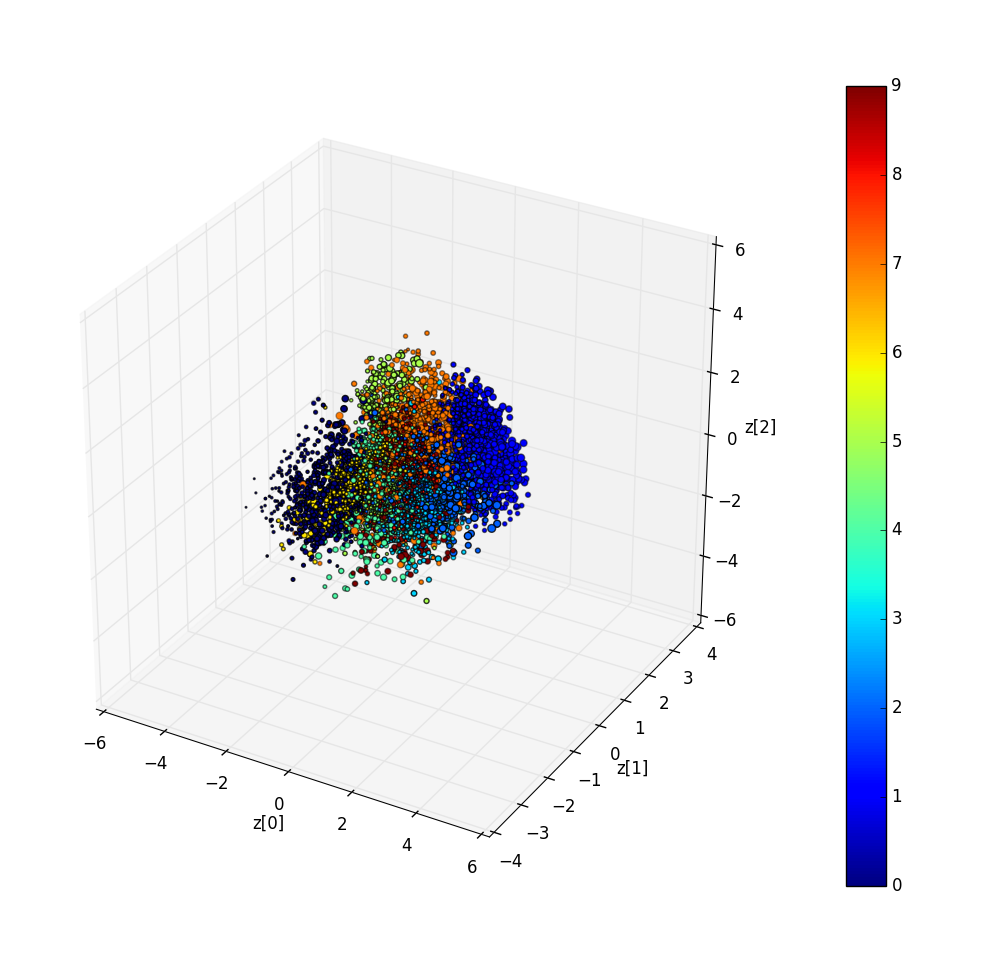}\vspace{-.3cm}\\ 
    \includegraphics[width=.09\textwidth]{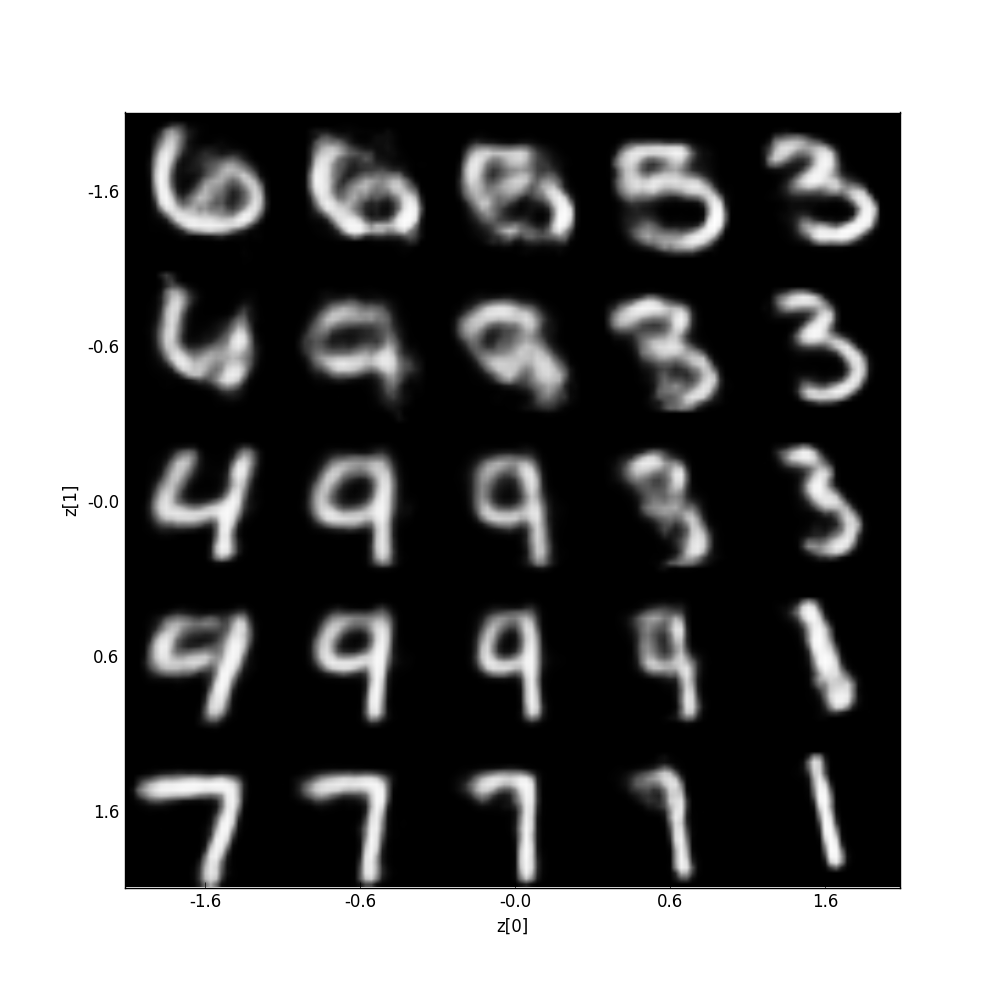}\!\!
    \includegraphics[width=.09\textwidth]{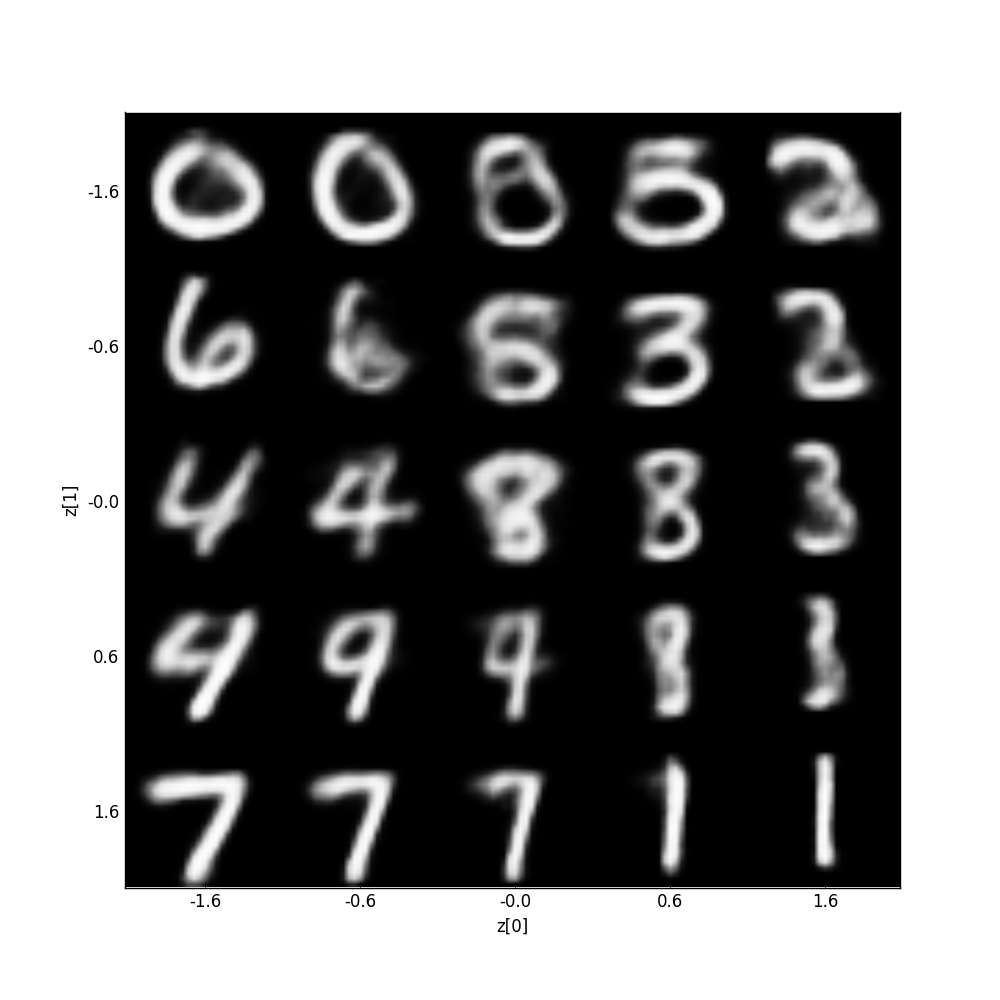}\!\!
    \includegraphics[width=.09\textwidth]{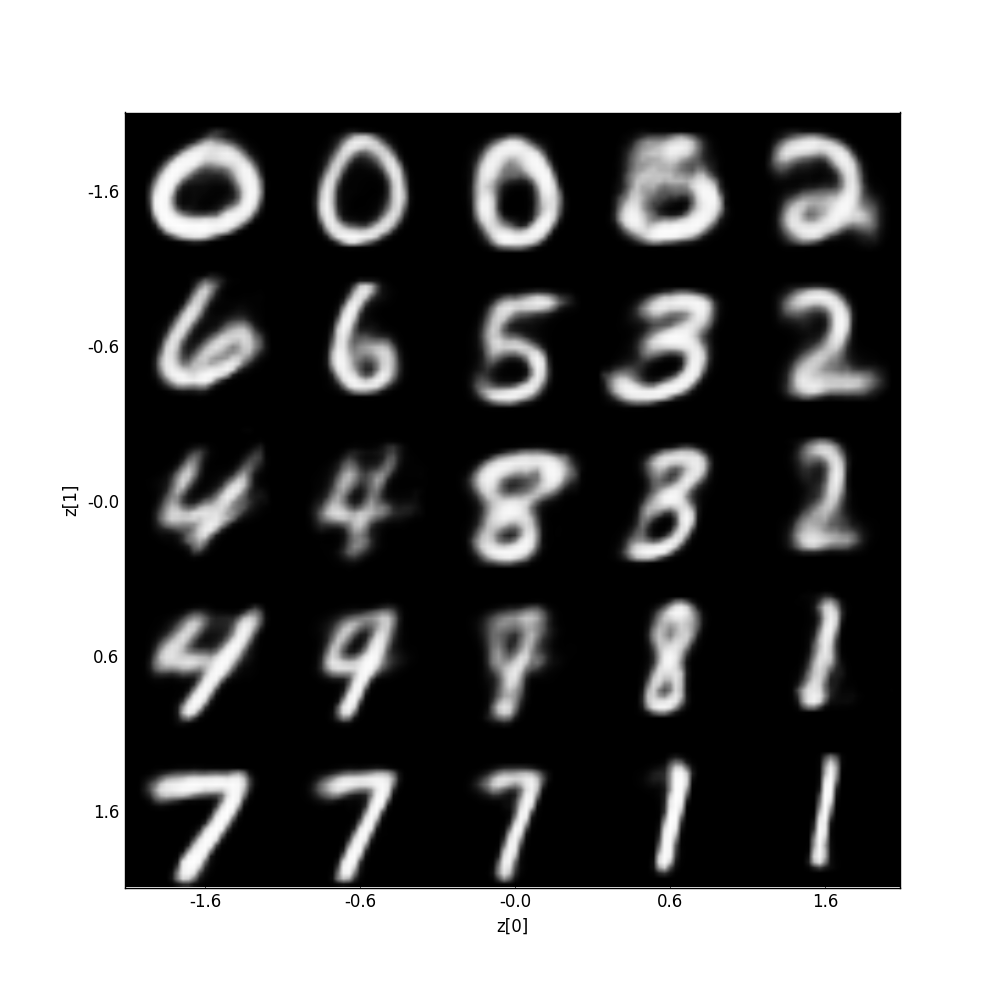}\!\!
    \includegraphics[width=.09\textwidth]{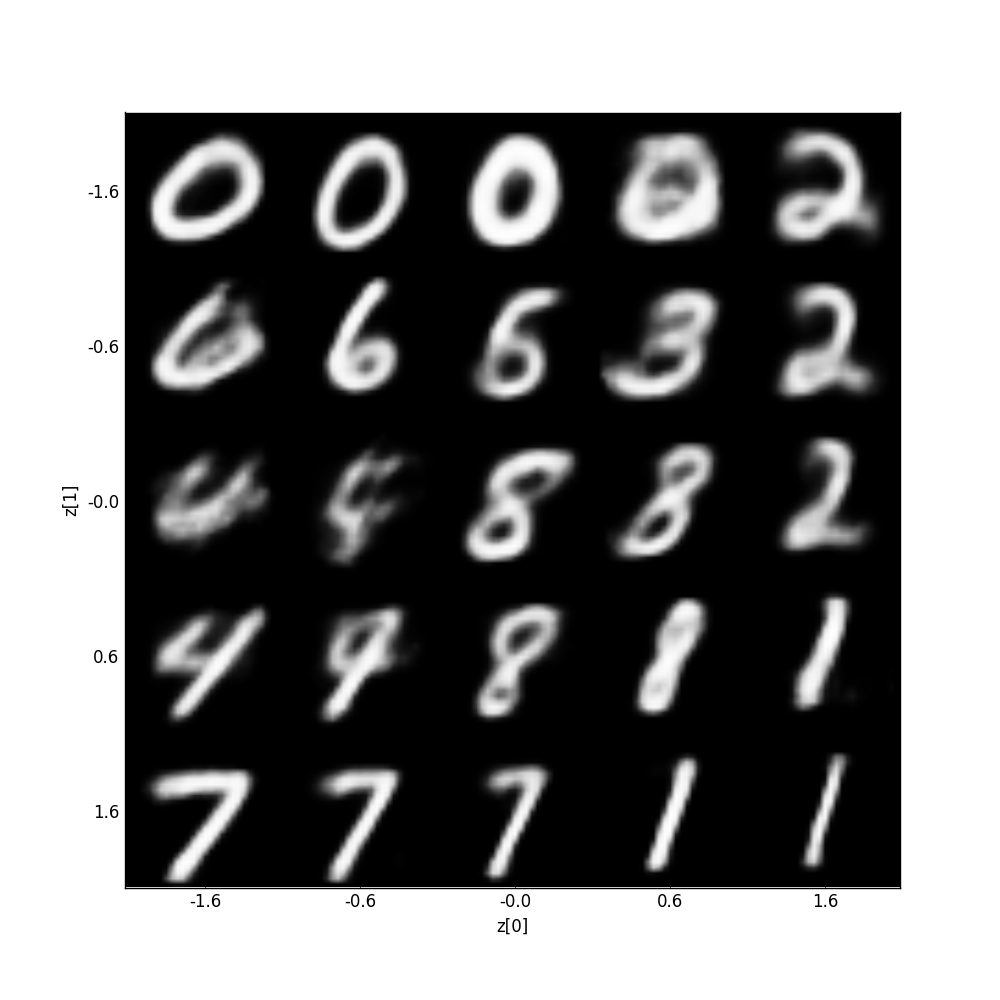}\!\!
    \includegraphics[width=.09\textwidth]{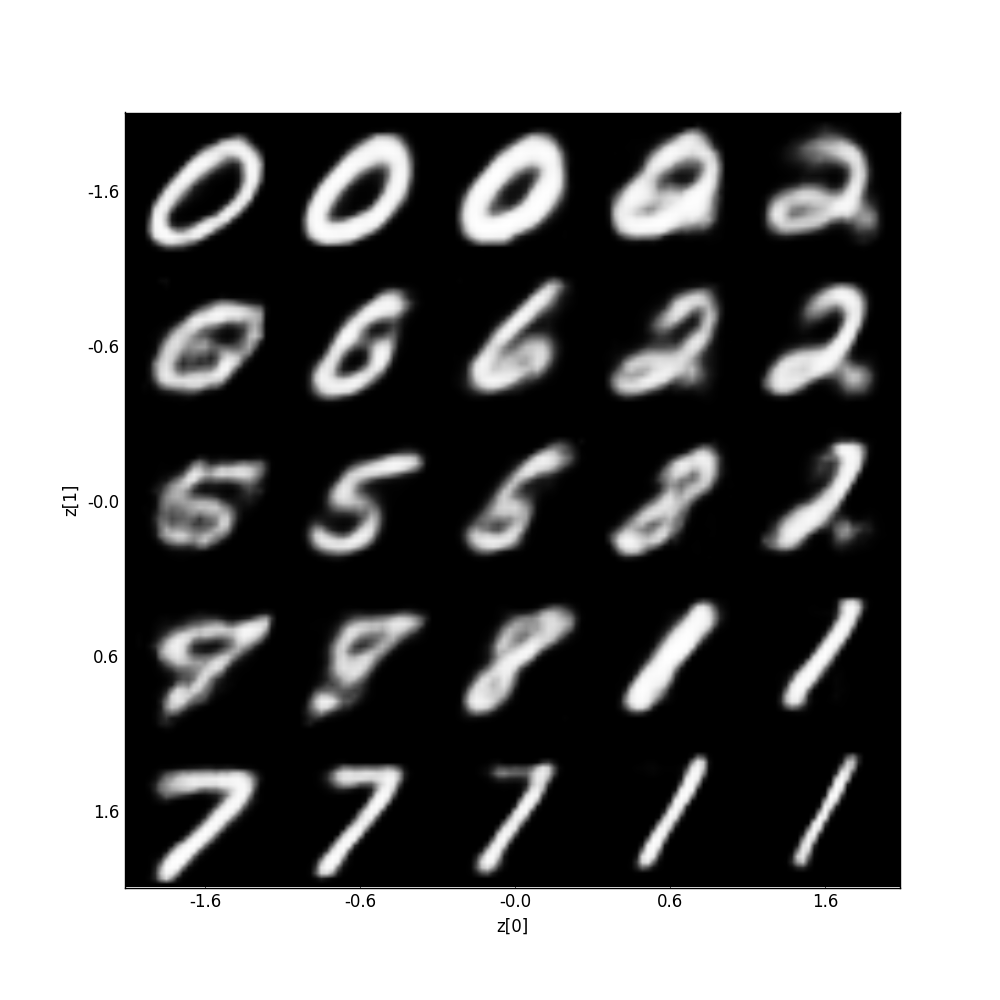}\\
    (A) VAE \\
    \includegraphics[width=.4\textwidth]{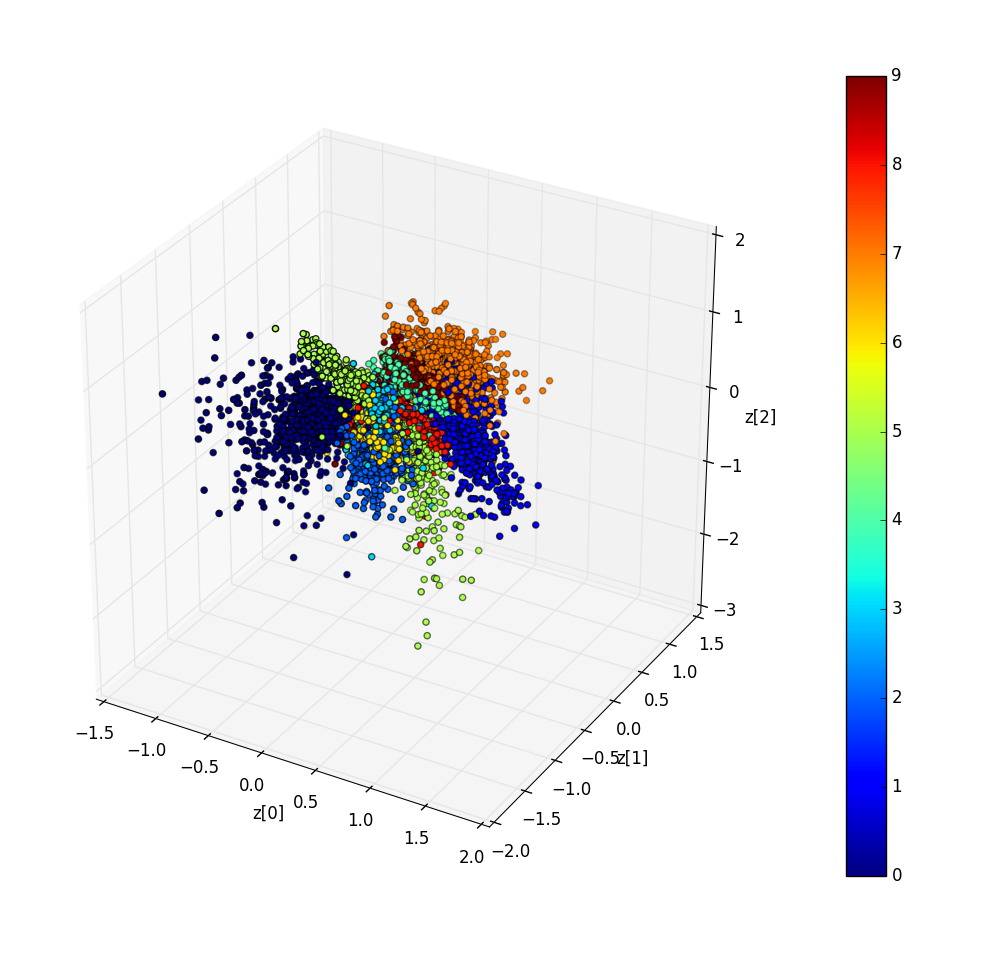}\vspace{-.3cm}\\\includegraphics[width=.09\textwidth]{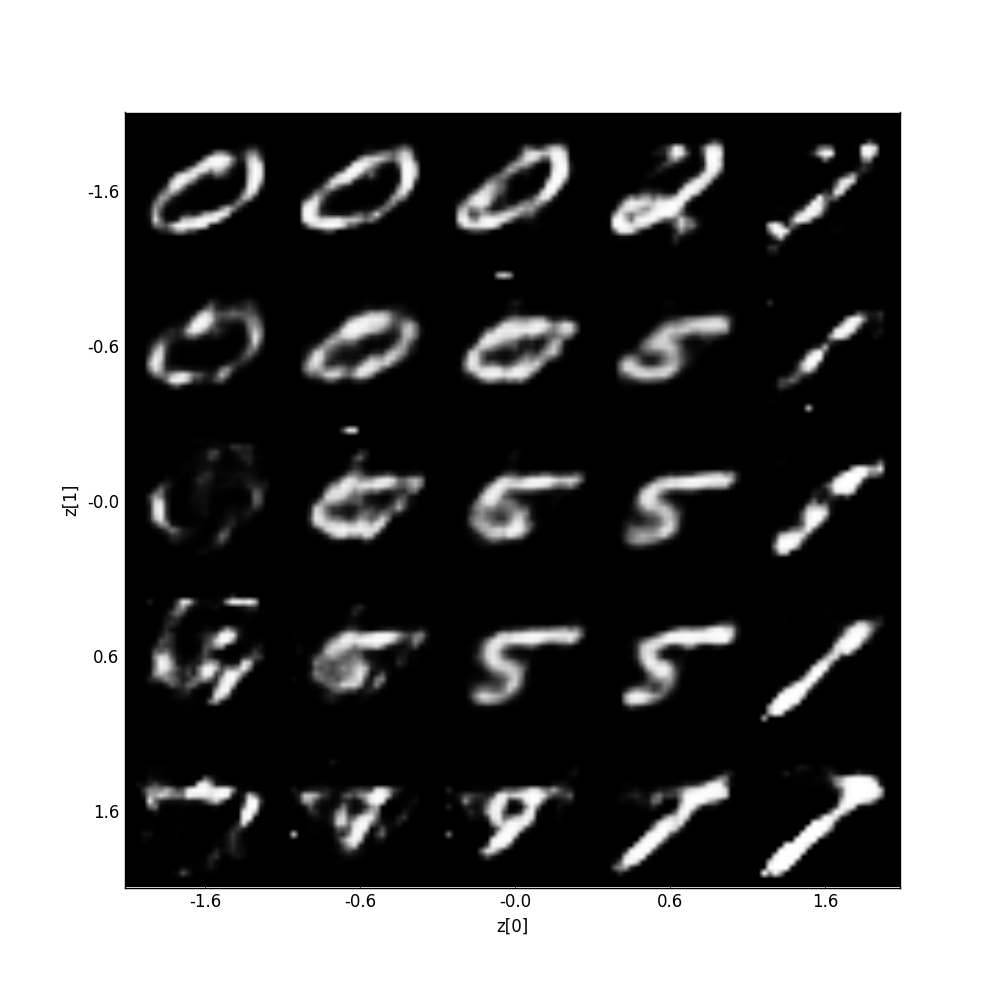}\!\! 
    \includegraphics[width=.09\textwidth]{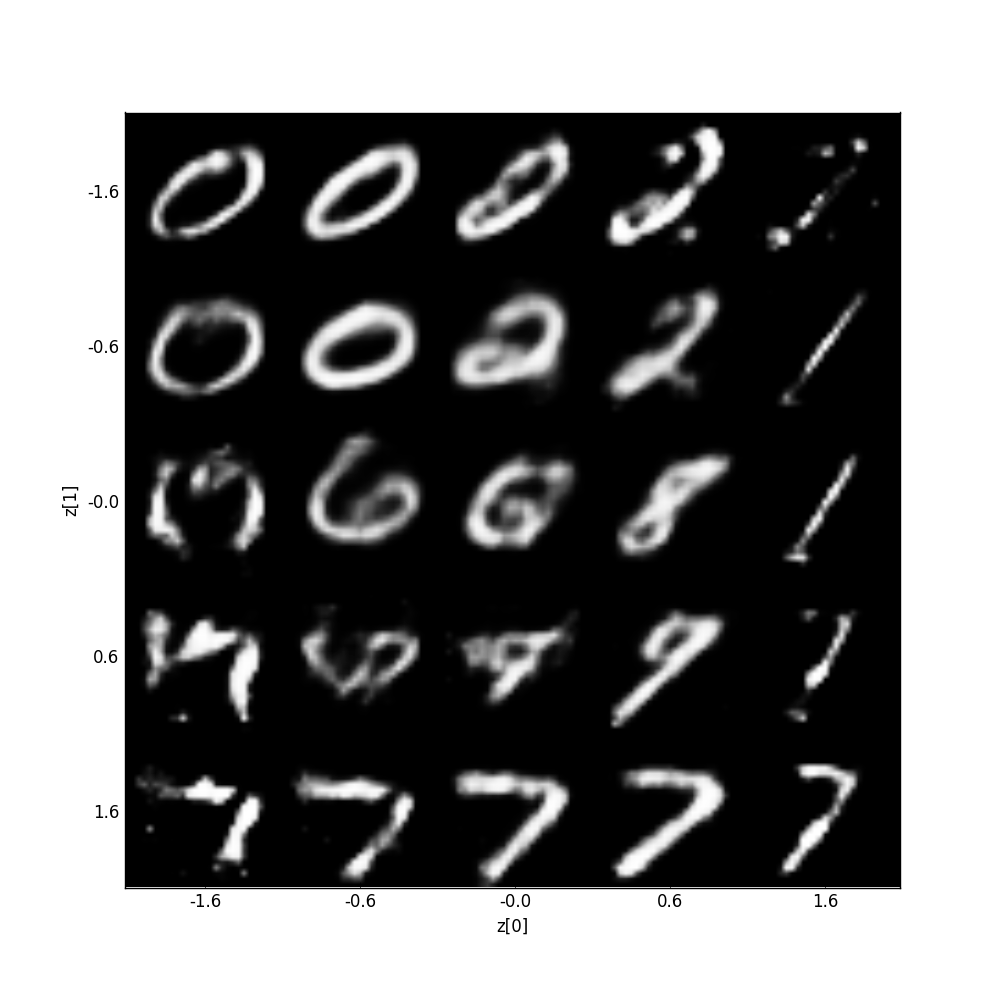}\!\!
    \includegraphics[width=.09\textwidth]{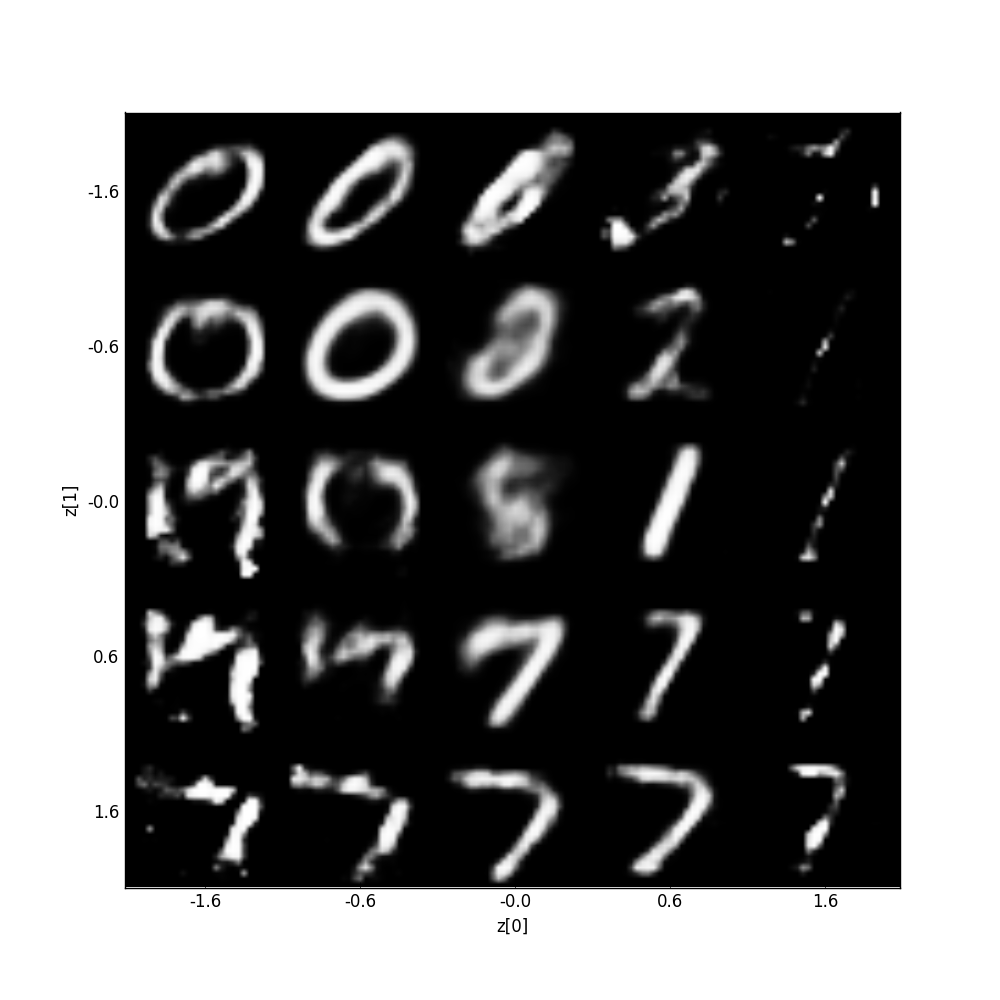}\!\!
    \includegraphics[width=.09\textwidth]{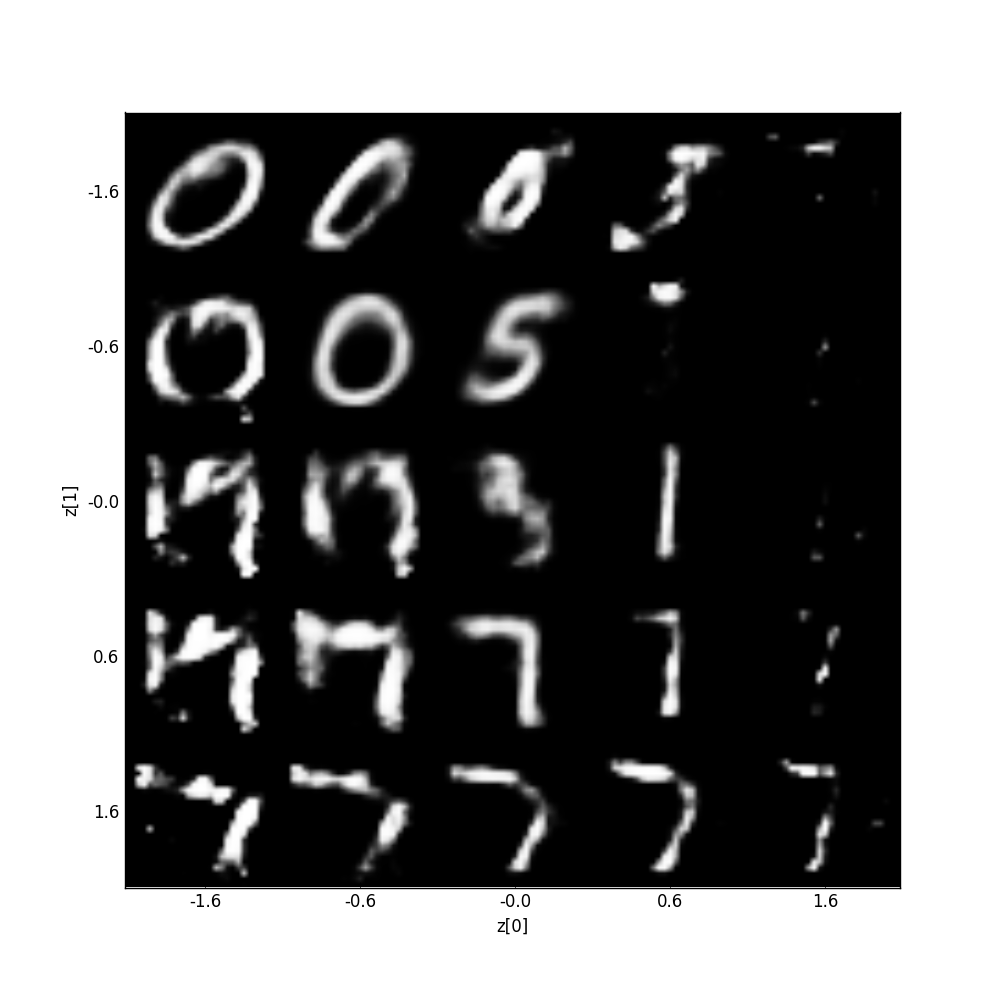}\!\!
    \includegraphics[width=.09\textwidth]{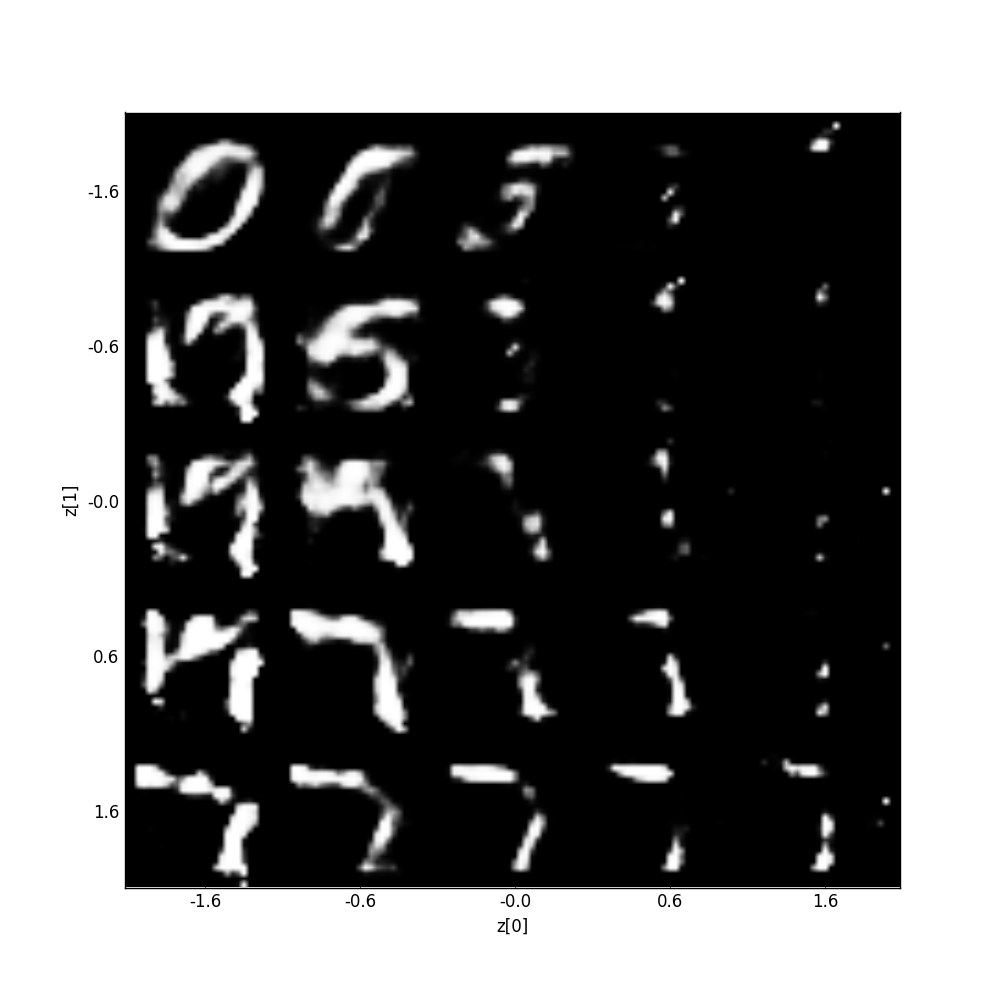}\\\\
    (B) without sampling at training time
    \end{tabular}
    \caption{Representation of 5000 MNIST digits in the latent space, and examples of
    generative sampling. Figure (A) is relative to a Variational Autoencoder with a dense 784-256-64-16-3 
    architecture, Figure (B) is the same, but without sampling at training time (we
    keep the quadratic penalty on latent variables). The disposition of encodings in (A) has much more spherical shape, resulting in 
    a drastic improvement in the quality of generated samples.
    }
    \label{fig:3D}
\end{figure}

In Figure~\ref{fig:3D}, we show the final distribution of 5000 MNIST digits in the 3-dimensional latent space with and without sampling during training (in the case without sampling we keep the quadratic penalty on $\mu_\theta(X)$). We also show the result of generative sampling from the latent space, organized in five horizontal slices of 25 points each. 

We may observe that sampling during training induces a much more regular disposition of points in the latent space.
In turn, this results in a drastic improvement in the quality of randomly generated images.


\subsection{Towards higher dimensions}
One could easily wonder how the variational approach scales to higher dimensions of the latent space. The fear is as usual related to the curse of dimensionality: in a high dimensional space encoding of data in the latent space will eventually be scattered away, and it is not evident that sampling during training will be enough to guarantee a good coverage in view of generation.

\begin{figure}
    \centering
    \includegraphics[width=.46\textwidth]{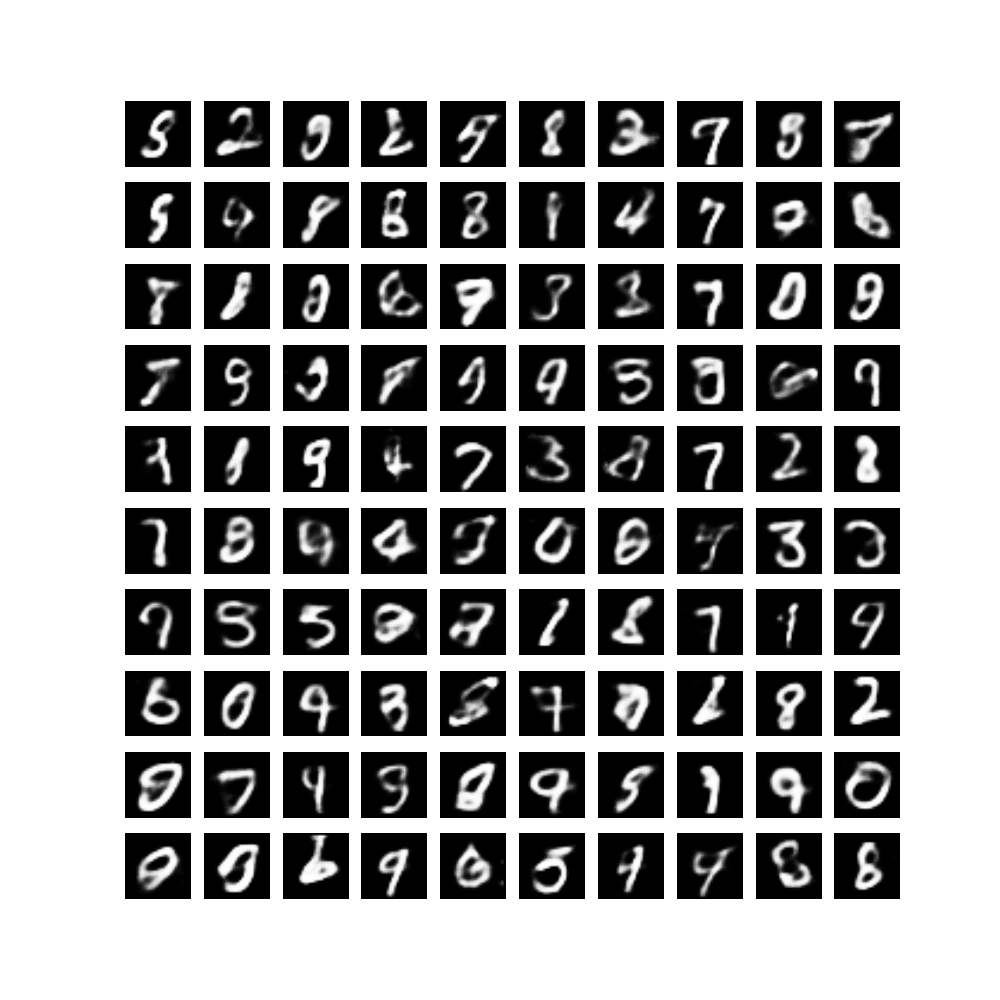}\vspace{-.4cm}
    \caption{Sampling from a latent spare of dimension 16 using a dense 784-256-32-24-16 VAE.}
    \label{fig:sampling16}
\end{figure}

In Figures~\ref{fig:sampling16} we show the result of generative sampling from a latent space of dimension 16, using a 784-256-32-24-16 dense VAE architecture. 
The generated digits are not bad, confirming a substantial stability of Variational Autoencoders to a growing dimension of the latent space. 
However, we observe a different and much more intriguing phenomenon: the internal representation is getting
{\em sparse}. 

\begin{figure}
    \centering
    \includegraphics[width=.5\textwidth]{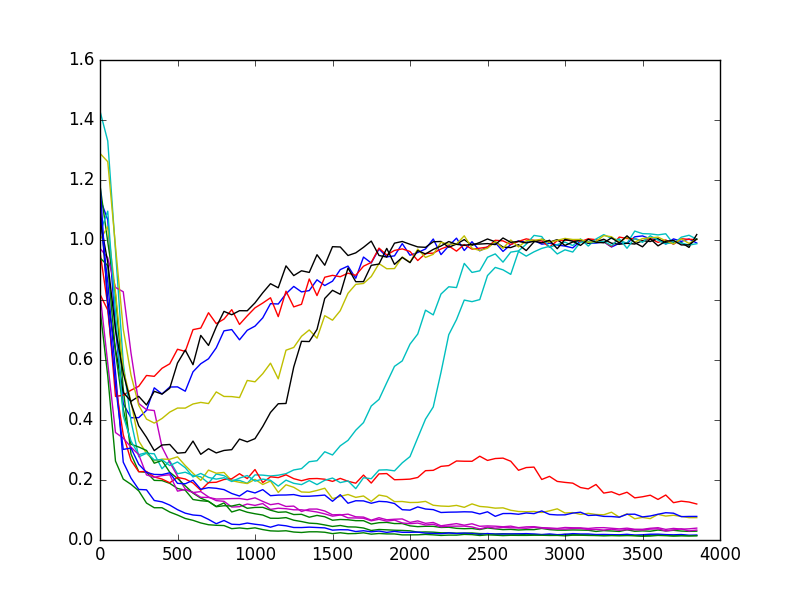}
    \caption{Evolution of the variance along training (16 variables, MNIST case). On the x-axis we have numbers of minibatches, each one of size 128.}
    \label{fig:evolution}
\end{figure}

In Figure~\ref{fig:evolution} we show the evolution during a typical
training of the variance of latent variables in  a space of dimension 16.
Table~\ref{tab:sparsity_mnist} provides relevant statistics for each latent variable
at the end of training, computed over the full dataset: the mean of its variance (that we expect to be around 1, since it should be normally distributed), 
and the mean of the computed variance $\sigma_\theta^2(X)$ (that we expect to be a small value, close to 0). The mean value is around 0 as expected, and we do not report it.
\begin{table}[ht]
    \centering
    \begin{tabular}{|c|c|c|}
    \hline
    no. & variance & mean($\sigma_\theta^2(X))$\\\hline
    \cellcolor{red!25}0 & \cellcolor{red!25}8.847272e-05 &\cellcolor{red!25} 0.9802212 \\\hline
    \cellcolor{red!25}1 & \cellcolor{red!25}0.00011756 & \cellcolor{red!25}0.99551463 \\\hline
    \cellcolor{red!25}2 & \cellcolor{red!25}6.665453e-05 & \cellcolor{red!25}0.98517334 \\\hline
    3 & 0.97417927 & 0.008741336 \\\hline
    4 & 0.99131817 & 0.006186147 \\\hline
    5 & 1.0012343 & 0.010142518 \\\hline
    6 & 0.94563377 & 0.057169348 \\\hline
    \cellcolor{red!25}7 & \cellcolor{red!25}0.00015841 & \cellcolor{red!25}0.98205334 \\\hline
    8 & 0.94694275 & 0.033207607\\\hline
    \cellcolor{red!25}9 & \cellcolor{red!25}0.00014789 & \cellcolor{red!25}0.98505586 \\\hline
   10 & 1.0040375 & 0.018151345\\\hline
   11 & 0.98543876 & 0.023995731 \\\hline
   \cellcolor{red!25}12 & \cellcolor{red!25}0.000107441 &  \cellcolor{red!25}0.9829797 \\\hline
   \cellcolor{red!25}13 & \cellcolor{red!25}4.5068125e-05 & \cellcolor{red!25}0.998983 \\\hline
   \cellcolor{red!25}14 & \cellcolor{red!25}0.00010853 & \cellcolor{red!25}0.9604088 \\\hline
   15 & 0.9886378 & 0.044405878 \\\hline
    \end{tabular}
    \caption{Inactive variables in the VAE for generating MNIST digits (dense case)}
    \label{tab:sparsity_mnist}
\end{table}

\begin{figure}[h!]
    \centering
    \includegraphics[width=.5\textwidth]{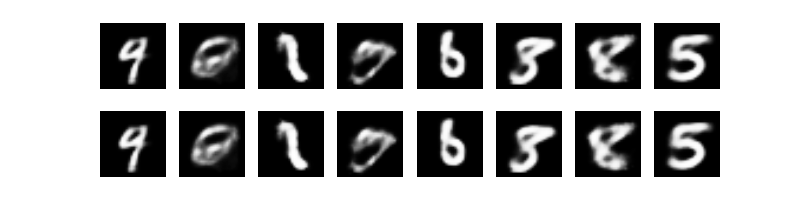}
    \caption{Upper line: digits generated from a vector of 16 normally sampled latent variables.
    Lower line: digits generated after "red" variables have been zeroed-out: these latent variables
    are completely neglected by the generator.}
    \label{fig:red}
\end{figure}
All variables highlighted in red have an anomalous behavior: their variance is very low (in practice,
they always have value 0), while the variance $\sigma_\theta^2(X)$ computed by the network 
is around $1$ for each $X$. 
Only 8 latent variables out of 16 are in use: the other ones (we call them {\em inactive}) are completely ignored by the
generator. 
For instance, in Figure~\ref{fig:red} we show a few digits randomly generated from
Gaussian sampling in the latent space (upper line) 
and the result of generation when inactive latent variables have been zeroed-out (lower line): they
are indistinguishable.

\subsection{Convolutional Case}\label{sec:conv_mnist}
With convolutional networks, sparsity is less evident. 
We tested a relatively sophisticated network, whose structure is summarized in Figure~\ref{fig:conv_arch}.
\begin{figure}[ht]
\begin{scriptsize}
\begin{verbatim}
  Layer (type)          Output Shape           Params                  
  ===================================================
  InputLayer            (None, 28, 28, 1)      0      
  ___________________________________________________
  conv2d 3x3            (None, 14, 14, 16)     160                    
  ___________________________________________________
  BatchNormalization    (None, 14, 14, 16)     64                    
  ___________________________________________________
  RELU                  (None, 14, 14, 16)     0          
  ___________________________________________________
  conv2d 3x3            (None, 14, 14, 32)     4640              
  ___________________________________________________
  conv2d 3x3            (None, 7, 7, 32)       9248                  
  ___________________________________________________
  BatchNormalization    (None, 7, 7, 32)       128                       
  ___________________________________________________
  RELU                  (None, 7, 7, 32)       0            
  ___________________________________________________
  conv2d 3x3            (None, 4, 4, 32)       9248              
  ___________________________________________________
  conv2d_3x3            (None, 4, 4, 32)       9248                   
  ___________________________________________________
  BatchNormalization    (None, 4, 4, 32)       128                      
  ___________________________________________________
  RELU                  (None, 4, 4, 32)       0      
  ___________________________________________________
  conv2d 3x3            (None, 2, 2, 32)       4128                
  ___________________________________________________
  conv2d 3x3            (None, 2, 2, 32)       4128                   
  ___________________________________________________
  BatchNormalization    (None, 2, 2, 32)       128              
  ___________________________________________________
  RELU                  (None, 2, 2, 32)       0    
  ___________________________________________________
  conv2d 3x3            (None, 1, 1, 32)       4128               
  ___________________________________________________
  conv2d 1x1            (None, 1, 1, 16)       528                      
  ___________________________________________________
  conv2d_1x1            (None, 1, 1, 16)       528          
\end{verbatim}
\end{scriptsize}
\caption{Architecture of the convolutional encoder. The two final layers compute mean and variance for
16 latent variables.
The decoder is symmetric, using transposed convolutions.}\label{fig:conv_arch}
\end{figure}

\begin{figure}[h!]
    \centering
    \includegraphics[width=.46\textwidth]{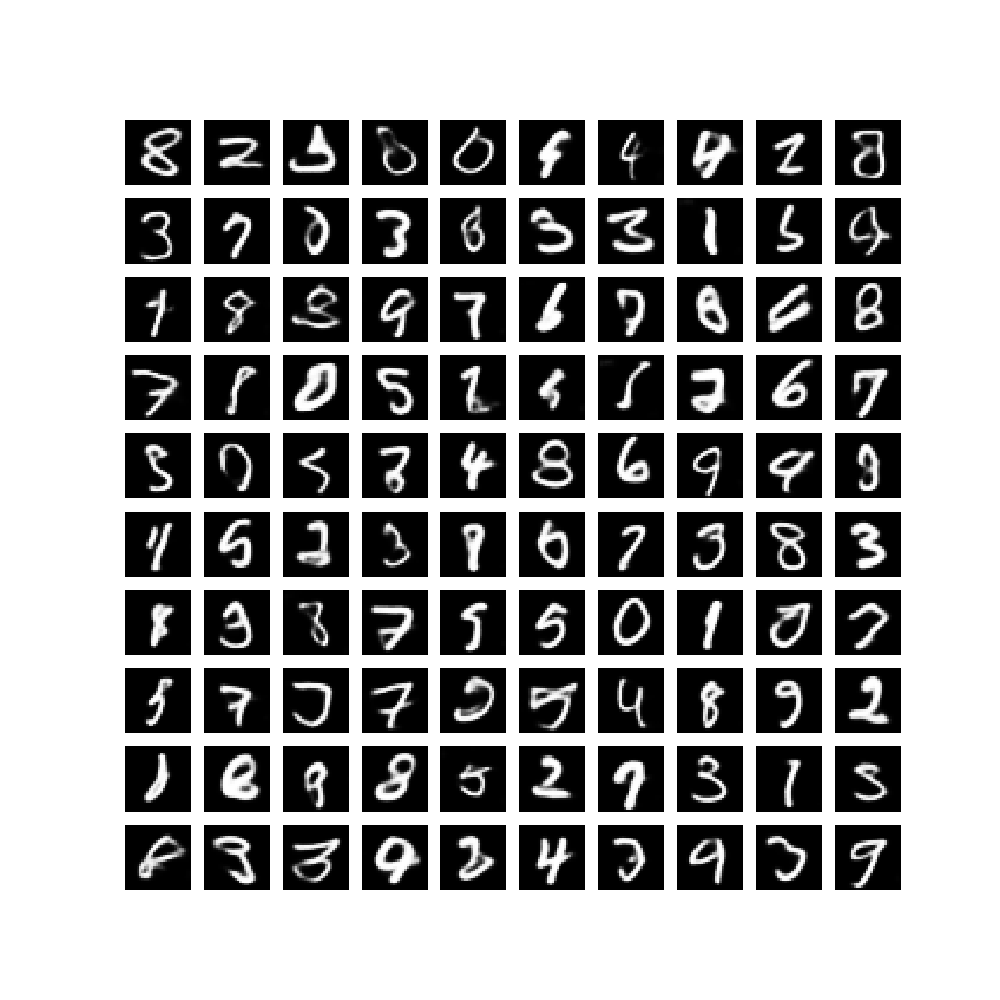}\vspace{-.5cm}
    \caption{Generation of MNIST digits via a convolutional network}
    \label{fig:gen_digits_conv}
\end{figure}

The previous network is able to produce excellent generative results
(see Figure~\ref{fig:gen_digits_conv}).

As for sparsity, 3 of the 16 latent variables are zeroed out. Having less sparsity seems to suggest that convolutional 
networks make a better exploitation of latent variables, typically resulting in a more precise reconstruction and improved generative sampling. 
This is likely due to the fact that latent
variables encode information corresponding to different portions of the input space, and are less likely to become useless for the generator.

\section{Generating tiles}
We repeat the experiment on a different generative problem, consisting 
in generating images containing a given number of small white
tiles on a black background. 
The number of expected tiles should depend on the kind of
training images provided to the autoencoder. In this case, we fed as training data
images containing a number of tiles ranging between 1 and 3. All tiles have dimension 
$4\times 4$ and the whole image is $28\times 28$, similarly to MINST.

We start addressing the problem with the a dense network with 
dimensions 784-512-256-64-32-16. 

\begin{figure}[hb]
    \centering
    \includegraphics[width=.5\textwidth]{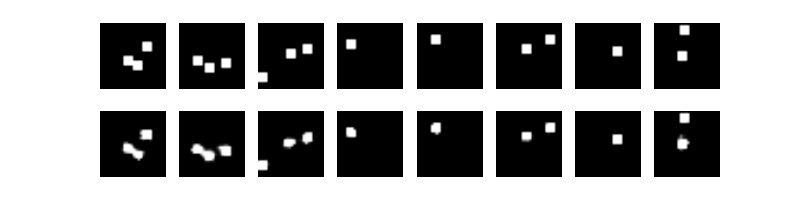}\vspace{-.2cm}
    \caption{Input images of the "counting tiles" problem (upper line) and corresponding reconstructions
    (lower line).}
    \label{fig:reconstruct}
\end{figure}

\begin{figure}[ht]
    \centering
    \includegraphics[width=.48\textwidth]{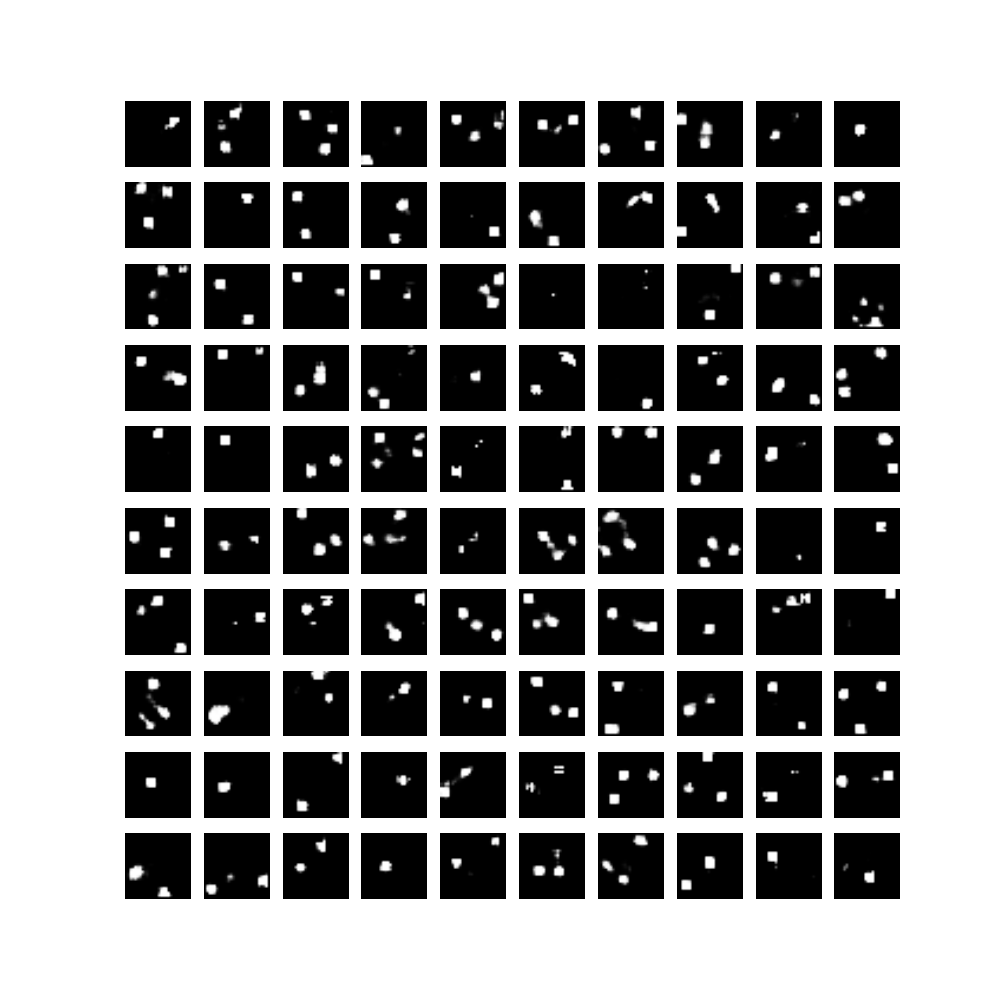}\vspace{-.6cm}
    \caption{Random generation of images with tiles from the latent space}
    \label{fig:sampling_tiles}
\end{figure}

In Figure~\ref{fig:reconstruct} we give some example of input images (upper row) and their corresponding reconstructions
(bottom row).
The mediocre reconstruction quality testifies that this is a
much more complex problem than MNIST. In particular, generative sampling is quite problematic (see Figure~\ref{fig:sampling_tiles})

Again, we have the same sparsity phenomenon already observed in the MNIST case: a large number of latent variables is inactive (see Table~\ref{tab:sparse_tiles}):
\begin{table}[ht]
    \centering
    \begin{tabular}{|c|c|c|}
    \hline
    no. & variance & mean($\sigma_\theta^2(X))$\\\hline
    0 & 0.89744579 & 0.08080574 \\
    \cellcolor{red!25}1 & \cellcolor{red!25}5.1123271e-05 &\cellcolor{red!25}  0.98192715\\\hline
    \cellcolor{red!25}2 & \cellcolor{red!25}0.00013507& \cellcolor{red!25} 0.99159979 \\\hline
    3 & 0.99963027 & 0.016493475 \\\hline
    \cellcolor{red!25}4 & \cellcolor{red!25}6.6830005e-05& \cellcolor{red!25} 1.01567184 \\\hline
    5 & 0.96189236 &  0.053041528 \\\hline
    6 & 1.01692736 &  0.012168014 \\\hline
    7 & 0.99424797 &  0.037749815 \\\hline
    \cellcolor{red!25}8 & \cellcolor{red!25} 0.00011436 &  \cellcolor{red!25}0.96450048\\\hline
    \cellcolor{red!25}9 & \cellcolor{red!25} 3.2284329e-05 & \cellcolor{red!25}0.97153729\\\hline
    \cellcolor{red!25}10 & \cellcolor{red!25}7.3369141e-05&  \cellcolor{red!25}1.01612401\\\hline
   11 & 0.91368156 & 0.086443416 \\\hline
   12 & 0.79746723175 & 0.23826576\\\hline
   \cellcolor{red!25}13 & \cellcolor{red!25}7.9485260e-05& \cellcolor{red!25} 0.9702732\\\hline
   14 & 0.92481815 &  0.089715622\\\hline
   \cellcolor{red!25}15 & \cellcolor{red!25}4.3311214e-05 & \cellcolor{red!25} 0.95554572\\\hline
    \end{tabular}
    \caption{Inactive variables in the VAE for generating tiles (dense case)}
    \label{tab:sparse_tiles}
\end{table}

\subsection{Convolutional Case}
\begin{figure}[ht]
    \centering
    \includegraphics[width=.4\textwidth]{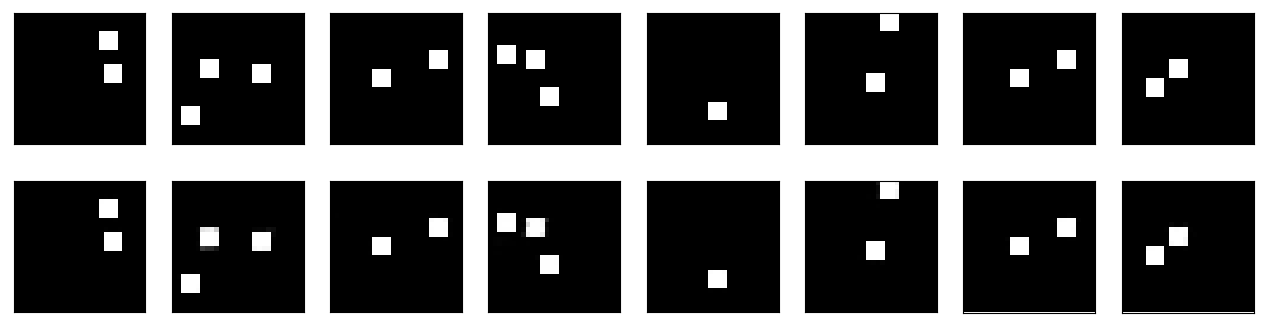}
    \caption{Input images of the "counting tiles" problem (upper line) and corresponding reconstructions
    (lower line) with a convolutional network.}
    \label{fig:reconstruct_conv}
\end{figure}

We tested the same architecture of Section~\ref{sec:conv_mnist}. In this case,  
reconstruction is excellent (Figure~\ref{fig:reconstruct_conv}).

\begin{figure}[h!]
    \centering\vspace{-.5cm}
    \includegraphics[width=.48\textwidth]{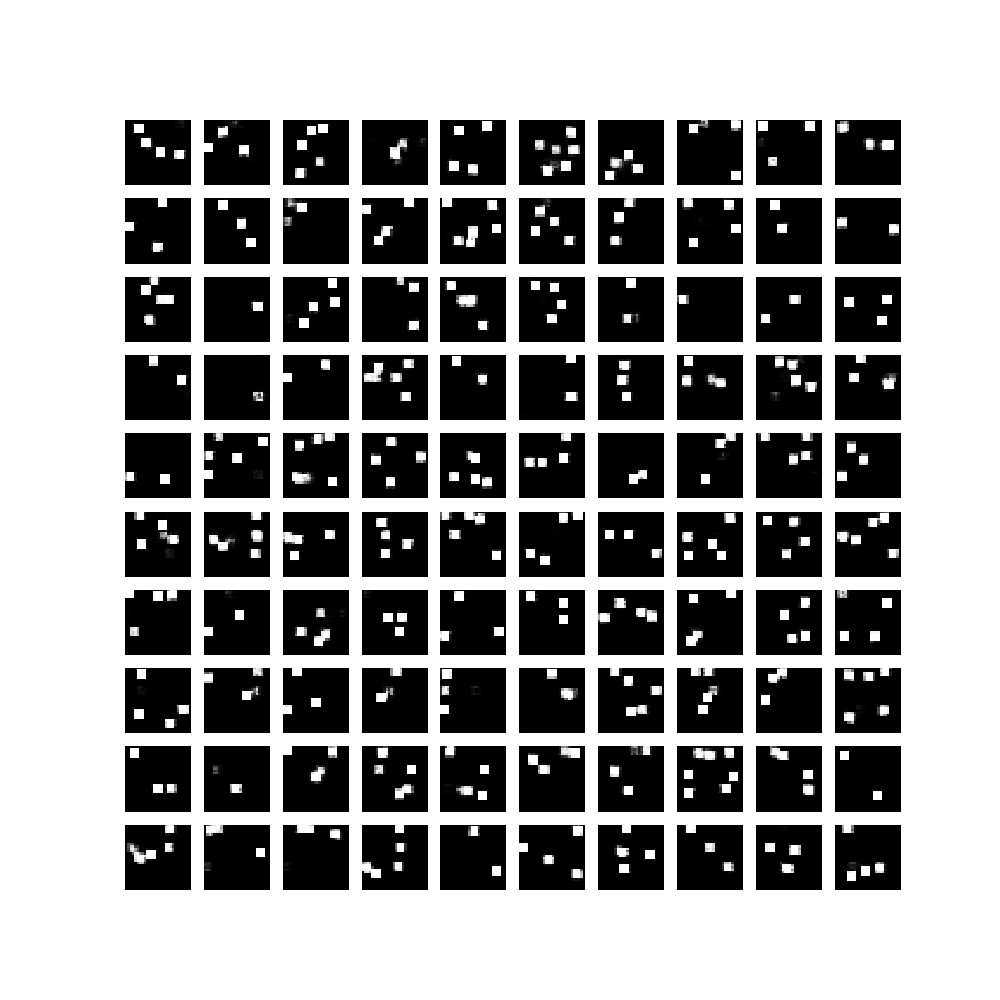}\vspace{-.6cm}
    \caption{Random generation of images with tiles from the latent space with a convolutional network.}
    \label{fig:sampling_tiles_conv}
\end{figure}

Generative sampling has improved too (see Figure~\ref{fig:sampling_tiles_conv}), but for an annoying counting problem: we expected to generate images with {\em at most three tiles}, while they frequently 
contain
a much larger number of them\footnote{This counting issue was the original motivation for our interest in this generative problem.}.

From the point of view of sparsity, 4 latent variables out
of 16 result to be inactive.

\section{Kullback-Leibler divergence and sparsity}\label{sec:sparsity}
Let us first of all observe that trying to compute relevant statistics for the posterior 
distribution $Q(z|X)$ of latent variables without some kind of regularization constraint does 
not make much sense. As a matter of fact, 
given a network with mean $\mu_z(X)$ and variance $\sigma^2_z(X)$ we can easily build
another one having precisely the same behaviour by scaling mean and standard deviation by some
constant $\gamma$ (for all data, uniformly), and then downscaling the generated samples 
in the next layer of the network. This kind of linear transformations are easily
performed by any neural network (it is the same reason why it does not make much sense 
to add a batch-normalization layer \cite{BatchNorm} before a linear layer). 

\begin{figure}[h!]
    \centering\vspace{-.3cm}
    \includegraphics[width=.48\textwidth]{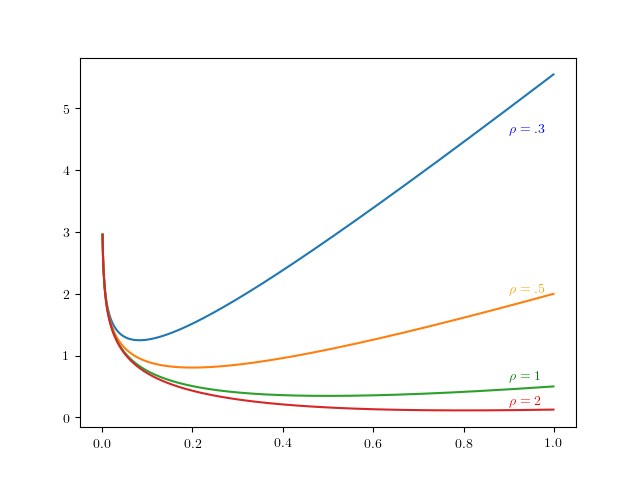}\vspace{-.6cm}
    \caption{KL-divergence for different values of $\rho$: observe the strong minimum for small values of
    $\rho$.}
    \label{fig:KL-rho}
\end{figure}

Let's see how the KL-divergence helps to choose a solution. In the following, we suppose to work 
on a specific latent variable $z$. Starting from the assumption that
for a network it is easy to keep a fixed ratio $\rho^2(X) = \frac{\sigma^2(X)}{\mu^2(X)}$
we can push this value in the closed form of the Kullback-Leibler divergence 
(see Equation~\ref{eq:closed-form}), getting the following expression 
\begin{equation}\label{eq:closed-rho}
\begin{array}{l}
  \frac{1}{2}(\sigma^2(X)\frac{1 + \rho^2(X)}{\rho^2(X)}-log(\sigma^2(X)) -1)
\end{array}
\end{equation}
In Figure~\ref{fig:KL-rho} we plot the previous function in terms of the variance, for a few given
values of $\rho$. 
It is easy to derive that we have a miminum for 
\begin{equation}\label{eq:minimum}
    \sigma^2(X) = \frac{\rho^2(X)}{1+\rho^2(X)}
\end{equation}
that is close to 0 when $\rho$ is
small, and close to $1$ when $\rho$ is high. Of course $\rho$ depends on $X$, while the rescaling operation must be uniform,
still the network will have a propensity to synthesize standard variations close to $0 <= \frac{\rho^2(X)}{1+\rho^2(X)} < 1$ (below we shall average on all X).

Substituting the definition of $\rho^2(X)$ in equation \ref{eq:minimum}, 
we expect to reach a minimum
when $\sigma^2(X) = \frac{\sigma^2(X)}{\mu^2(X)}\frac{\mu^2(X)}{\mu^2(X)+\sigma^2(X)}$,
that, by trivial computations, implies the following simple stationary condition:
\begin{equation}\label{eq:stationarity}
    \sigma^2(X) + \mu^2(X) = 1
\end{equation}

Let us now average together the KL components for all data X:
\begin{equation}\label{eq:closed-average}
\frac{1}{N}\sum_X \frac{1}{2}(\mu(X)^2 + \sigma^2(X)-log(\sigma^2(X)) -1)
\end{equation}
We use the notation $\widehat{f(X)}$ to abbreviate the average $\frac{1}{N}\sum_X f(X)$ of $f(X)$ on all data $X$. 
The ratio $\rho^2 = \frac{\widehat{\sigma^2(X)}}{\widehat{\mu^2(X)}}$ can really (and easily) be kept constant by the net. Let us also observe that, assuming the mean of the latent variable to be 0, $\widehat{\mu^2(X)}$ is just the (global) variance $\sigma^2$ of the latent variable. Pushing
$\rho^2$ in the previous equation, we get
\[
  \frac{1}{2}(\widehat{\sigma^2(X)}\frac{1 + \rho^2}{\rho^2}-\widehat{log(\sigma^2(X))} -1)
\]
The average of the logarithms $\widehat{log(\sigma^2(X))}$ is the logarithm of the geometric
mean of the variances. If we replace the geometric mean with an arithmetic mean, we get an
expression essentially equivalent to equation \ref{eq:closed-rho}, namely
\begin{equation}\label{eq:rho-average}
 \frac{1}{2}(\widehat{\sigma^2(X)}\frac{1 + \rho^2}{\rho^2}-log(\widehat{\sigma^2(X)}) -1)
\end{equation}
that has a minimum when $
    \widehat{\sigma^2(X)} = \frac{\rho^2}{1+\rho^2}
$, that implies
\begin{equation}\label{eq:major}
    \widehat{\sigma^2(X)} + \widehat{\mu^2(X)} = 1
\end{equation}
or simply, 
\begin{equation}\label{eq:major-sigma}
    \widehat{\sigma^2(X)} + \sigma^2 = 1
\end{equation}
where we replaced $\widehat{\mu^2(X)}$ with the variance $\sigma^2$ of the latent variable in view of the consideration above.

Condition \ref{eq:major} can be experimentally verified. We did it in several experiments, and it always proved to be quite accurate, provided the neural network was sufficiently trained. 
As an example, 
for the data in Table \ref{tab:sparsity_mnist} and Table \ref{tab:sparse_tiles} the average {\em sum} of $\widehat{\sigma^2(X)}$ and
$\sigma^2$ (the two cells in each row) is $0.997$ with a variance of $0.00048$! Let us remark that condition \ref{eq:major-sigma} holds both for active and inactive variables, and not just for the cases when values are close to 0 or 1; for instance, observe the case of variable 12 in table , which has a global variance around 0.797 and a 
mean local variance $\widehat{\sigma^2(X)}$ around 0.238, almost complementary to 1.

\subsection{Sparsity}
Let us consider again the loglikelihood for data X.
\[log(P(X)) \approx \EX_{z\sim Q(z|X)}log(P(X|z) - KL(Q(z|X)||P(z))\]
If we replace the Kullback-Leibler component with some other regularization, 
for instance just considering a quadratic penalty on latent variables, the sparsity phenomenon
disappears. So, sparsity is tightly related to the Kullback-Leibler divergence and in particular to the part of the term trying to keep the variance close to 1, that is
\begin{equation}
  - \sigma^2_\theta(X)+log(\sigma^2_\theta(X)) + 1 \label{eq:sigma_loss}
\end{equation}
whose effect typically degrades the distinctive characteristics of the features.
It is also evident that if the generator ignores a latent variable, 
$P(X|z)$ will not depend on it and the loglikelihood is maximal when the distribution of
$Q(z|X)$ is equal to the prior distribution $P(z)$, that is just a normal distribution
with 0 mean and standard deviation 1. In other words, the generator is induced to learn
a value $\mu_\theta(X) = 0$, ans a value $\sigma_\theta(X)=1$; sampling has no
effect, since the sampled value for $z$ will just be ignored. 

During training, if a latent variable is of moderate interest for reconstructing the
input (in comparison with the other variables), the network will learn to give less 
importance to it; at the end, the Kullback-Leibler
divergence may prevail, pushing the mean towards 0 and the standard deviation towards 1.
This will make the latent variable even more noisy, in a vicious loop that will eventually
induce the network to completely ignore the latent variable.

\begin{figure}[ht]
    \centering
    \includegraphics[width=.5\textwidth]{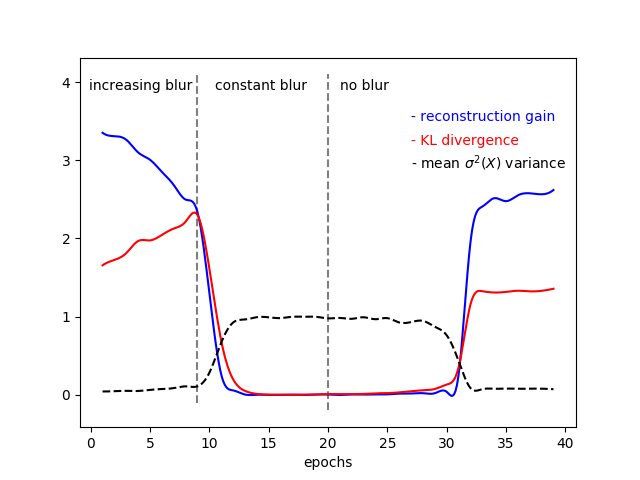}
    \caption{Evolution of reconstruction gain and KL-divergence of a latent variable 
    during training, acting on its quality by addition of Gaussian blur. We also show
    in the same picture the evolution of the variance, to compare their progress.}
    \label{fig:progressive}
\end{figure}

We can get some empirical evidence of the previous phenomenon by artificially deteriorating
the quality of a specific latent variable.
In Figure~\ref{fig:progressive}, we show the evolution during training of one of the active variables
of the variational autoencoder in Table~\ref{tab:sparsity_mnist} subject to a progressive
addition of Gaussian noise. During the experiment, we force the variables that were already 
inactive to remain so, otherwise the network would compensate the deterioration of a new
variable by revitalizing one of the dead ones.

In order to evaluate the contribution of the variable to the loss function we compute the difference between the reconstruction error when
the latent variable is zeroed out with respect to the case when it is normally taken into account; 
we call this information {\em reconstruction gain}.

After each increment of the Gaussian noise we repeat one epoch of training, to allow the
network to suitably reconfigure itself. In this particular case, the network reacts 
to the Gaussian noise by enlarging the mean values $\mu_x(X)$, in an attempt to escape
from the noisy region, but also jointly increasing the KL-divergence. At some point, the
reconstruction gain of the variable is becoming less than the KL-divergence; at this point
we stop incrementing the Gaussian blur. Here, we assist to the sparsity phenomenon: 
the KL-term is suddenly pushing variance towards 1 (due to equation \ref{eq:sigma_loss}), with the result of decreasing the
KL-divergence, but also causing a sudden and catastrophic collapse of the reconstruction 
gain of the latent variable. 

Contrarily to what is frequently believed, sparsity seems to be reversible, at some extent.
If we remove noise from the variable, as soon as the network is able to perceive a 
potentiality in it (that may take several epochs, as evident if Figure~\ref{fig:progressive}),
it will eventually make a suitable use of it. Of course, we should not expect to recover
the original information gain, since the network may have meanwhile learned a different 
repartition of roles among latent variables.

\section{A controversial issue}
The sparsity phenomenon in Variational Autoencoders is a controverial topic:
you can either stress the suboptimal use of the actual network capacity 
(overpruning), or its beneficial regularization effects. In this section we shall
rapidly survey on the recent research along this two directions.

\subsection{Overpruning}\label{sec:tackling}
The observation that working in latent spaces of sufficiently high-dimension,
Variational Autoencoders tend to neglect a large number of latent variables,
likely resulting in impoverished, suboptimal generative models,
was first made in \cite{BurdaGS15}. The term {\em overpruning} to
denote this phenomenon was introduced in \cite{overpruning17}, with a clearly
negative acception: an issue to be solved to improve VAE. 

The most typical approach to tackle this issue is that of using a parameter to trade off the contribution of the reconstruction error with respect to the Kullback-Leibler regularizer:

\begin{equation}
\begin{array}{l}
    - log(P(X)) \approx \\
    \hspace{.5cm}-\EX_{z\sim Q(z|X)}log(P(X|z) + \lambda KL(Q(z|X)||P(z))
\end{array}\label{eq:obj-lambda}
\end{equation}

The theoretical role of this $lambda$-parameter is not so evident; let us briefly discuss it. In the closed form of the traditional logloss fo VAE there are two parameters that seems to come out of the blue, and that may help to understand 
the $\lambda$. The first one is the variance of the prior distribution, that seems to be arbitrarily set to 1. However, as e.g. observed in \cite{tutorial-VAE}, a different variance for the prior may be easily compensated by the learned means $\mu(X)$ and variances $\sigma^2(X)$ for the posterior distribution $Q(z|X)$: in other words, the variance of the prior has essentially the role of fixing a unit of measure for the latent space. The second choice that looks arbitrary is the assumption that the distribution $P(X|z)$ has a normal shape around the decoder function $d_\theta(z)$: in fact, in this case, the variance of this distribution may strongly affect the resulting loss function, and could justify the introduction of a balancing $\lambda$ parameter.

Tuning down $\lambda$ reduces the number of inactive latent variable, but this may not result in an improved quality of generated samples: the network uses
the additional capacity to improve the quality of reconstruction, at the price
of having a less regular distribution in the latent space, that becomes harder to exploit by a random generator. 

More complex variants of the previous technique comprise an annealed optimization schedule for $\lambda$ \cite{Bowman15} or enforcing minimum KL contribution from subsets of latent units \cite{autoregressive16}. All these schemes require hand-tuning and, to cite 
\cite{overpruning17}, they easily risk to ``take away the principled regularization scheme that is built into VAE.''

A different way to tackle overpruning is that model-based, consisting in devising architectural modifications that may alleviate the problem. For instance, in \cite{epitomic2017} the authors propose a probabilistic generative model composed by a number of sparse variational autoencoders called epitoms that partially share their encoder-decoder architectures. The intuitive idea is that each data $X$ can be embedded 
into a small subspace $K_X$ of the latent space, specific to the given data.

Similarly, in \cite{Dieng2018AvoidingLV} the use of {\em skip-connections} is advocated as a possible technique to address over-pruning. 

While there is no doubt that particular network architectures show less
sparsity than others (see also the comparison we did in this article between dense and convolutional networks), in order to claim that the aforementioned
approaches are general techniques for {\em tackling over-pruning} it should 
be proved that they systematically lead to improved generative models across multiple architectures and many different data sets, that is a result still in want of confirmation.

\subsection{Regularization}\label{sec:regular}
Recently, there have been a few works trying to stress the beneficial effects
of the Kullback-Leibler component, and its essential role for generative purposes. 

An interesting perspective on the calibration between the reconstruction error and
the Kullback-Leibler regularizer is provided by $\beta$-VAE \cite{beta-vae17}
\cite{understanding-beta-vae18}. 
Formally, the shape of the objective function is the same 
of equation~\ref{eq:obj-lambda} (where the parameter $\lambda$ is renamed $\beta$), but
in this case the emphasis is in pushing $\beta$ to be {\em high}. 
This is reinforcing the 
sparsity effect of the Kullback-Leibler regularizer, inducing the model to learn more
disentangled features. The intuition is that the network should naturally learn
a representation of points in the latent space such that the ``confusion''
due to the Kullback-Leibler component is minimized: latent features should be
general, i.e. apply to a large number of data, and data should naturally cluster
according to them.
A metrics to measure the degree of disentanglement learned by the model is introduced in \cite{beta-vae17}, and it is used to provide experimental results confirming the beneficial effects of a strong regularization. In \cite{understanding-beta-vae18}, an interesting analogy between $\beta$-VAE and the Information Bottleneck is investigated.

In a different work \cite{RobustPCA}, it has been recently proved that 
a VAE with affine decoder is identical to a robust PCA model, able to decompose 
the dataset into a low-rank representation and a sparse noise. This is extended
to the nonlinear case in \cite{TwoStage}; in particular, it is proved that a VAE with infinite capacity can detect the manifold dimension and only use a minimal number of latent dimensions to represent the data, filling the redundant dimensions with white noise. In the same work the authors propose a quite interesting two stage approach, to
address the potential mismatch between the aggregate posterior $Q(z)$ and the prior $P(z)$: a second VAE is trained
to learn an accurate approximantion of $Q(z)$; samples from a Normal distribution are first used to generate samples
of $Q(z)$, and then fed to the actual generator of data points. In this way, it no longer matters that $P(z)$ and $Q(z)$ are not similar, since you can just sample from the latter using the second-stage VAE.  This approach does not
require additional hyperparameters or sensitive tuning, and produces high-quality samples, competitive 
with state-of-the-art GAN models, both in terms of FID score and visual quality.


\section{Conclusions}
In this article we discussed the interesting phenomenon of {\em sparsity} (aka over-pruning) in Variational
Autoencoders, induced by the Kullback-Leibler component of the objective function, and briefly surveyed some of the recent literature on the topic. Our point of view is 
slightly different from the most
traditional one, in the sense that maybe, as it is also suggested by other recent literature 
(see Section~\ref{sec:regular}), there is no issue to tackle: the Kullback-Leibler component has a beneficial self-regularizing effect, 
forcing the model to focus on the most important and disentangled features. This is 
precisely the reason why we prefer to talk of sparsity, instead of over-pruning. 
In particular, sparsity is
one of the few methodological guidelines that may be deployed in the architectural design of
Variational Autoencoders, suitably tuning the dimension of the latent space. If the resulting
network does not give satisfactory results, we should likely switch to
more sophisticated architectures, making a better exploitation of the latent space. 
If the reconstruction 
error is low but generation is bad, it is a clear indication of a mismatch between the aggregate posterior
$Q(z)$ and the prior $Q(z)$; in this case, a simple two-stage approach as described in \cite{TwoStage} 
might suffice.

Of the two terms composing the objective function of VAE, the weakest one looks the reconstruction error 
(traditionally dealt with a pixel-wise quadratic error), so it is a bit surprising that most of the research 
focus on the Kullback-Leibler regularization component.

\bibliographystyle{plain}
\bibliography{machine.bib,variational.bib}

\end{document}